\documentclass{article}

\usepackage{PRIMEarxiv}

\usepackage[utf8]{inputenc}
\usepackage[T1]{fontenc}
\usepackage[numbers]{natbib}
\usepackage{hyperref}
\usepackage{url}

\providecommand{\doi}[1]{\href{https://doi.org/#1}{\nolinkurl{#1}}}
\usepackage{booktabs}
\usepackage{amsfonts}
\usepackage{amsmath}
\usepackage{nicefrac}
\usepackage{microtype}
\usepackage{graphicx}
\usepackage{caption}
\usepackage{tikz}
\usepackage{xcolor}
\usepackage{verbatim}
\usepackage{cleveref}
\usepackage{fontawesome5}
\usepackage{orcidlink}

\graphicspath{{media/}}

\Crefname{equation}{Eq.}{Eqs.}
\Crefname{figure}{Fig.}{Figs.}
\Crefname{tabular}{Tab.}{Tabs.}

\title{QG-MIL: A Gated Transformer Aggregator for Domain-Agnostic Multiple Instance Learning in Medical Imaging}

\author{
\textbf{Luca Zedda}\textsuperscript{1}\,\orcidlink{0009-0001-8488-1612}\thanks{Equal contribution with Davide Antonio Mura. Co-corresponding authors: \{luca.zedda,davideantonio.mura\}@unica.it.}
\quad
\textbf{Davide Antonio Mura}\textsuperscript{1}\,\orcidlink{0009-0002-0701-9583}
\quad
{\mdseries Cecilia Di Ruberto}\textsuperscript{1}\,\orcidlink{0000-0003-4641-0307}
\\
{\mdseries Maurizio Atzori}\textsuperscript{1}\,\orcidlink{0000-0001-6112-7310}
\quad
{\mdseries Muhammed Furkan Dasdelen}\textsuperscript{2}\,\orcidlink{0000-0003-2251-2093}
\quad
{\mdseries Carsten Marr}\textsuperscript{2}\,\orcidlink{0000-0003-2154-4552}
\quad
{\mdseries Andrea Loddo}\textsuperscript{1}\,\orcidlink{0000-0002-6571-3816}
\\[0.8em]
{\mdseries \textsuperscript{1}Department of Mathematics and Computer Science, University of Cagliari, Cagliari, Italy}
\\
{\mdseries \textsuperscript{2}Institute of AI for Health, Helmholtz Munich, Neuherberg, Germany}
\\[0.4em]
{\mdseries \texttt{\{luca.zedda,davideantonio.mura\}@unica.it}}
}

\begin{document}

\maketitle

\begin{abstract}
    
Attention-based Multiple Instance Learning aggregators in medical imaging are prone to attention concentration, producing overconfident and unstable predictions. We introduce QG-MIL, a gated transformer aggregator that addresses this through four synergistic architectural components: RMSNorm-based pre-normalization, per-head QK normalization, fine-grained attention output gating, and SwiGLU-style feed-forward modules. Together, these design choices stabilize training and distribute attention more uniformly across instances without auxiliary losses, masking, or multi-stage regularization. We evaluate QG-MIL across six benchmarks spanning whole-slide pathology and cell-level hematology, covering two fundamentally different MIL scales. The best-performing QG-MIL variants outperform leading baselines on all six benchmarks, with an average improvement of +6.1 mean macro F1 points. Attention overlays and attention mass analysis confirm more distributed instance weighting. Ablation studies show that while individual components can match the full model on specific datasets, the QG-MIL design provides the most consistent cross-domain performance and tightest variance when compared to selected baselines. We release a configurable implementation to support reproducibility at: \\ https://github.com/unica-visual-intelligence-lab/QG-MIL
\keywords{Multiple Instance Learning \and Weakly Supervised Classification\and Gated Transformer\and Digital Pathology\and Hematology}

\end{abstract}
\begin{center}
\vspace{0.5em}
\noindent
\faGithub\ \textbf{Code:} \url{https://github.com/unica-visual-intelligence-lab/QG-MIL}
\quad
\end{center}

\section{Introduction}
\label{sec:intro}
Multiple Instance Learning (MIL) has become a core paradigm for
computational pathology and medical imaging tasks where slide- or patient-level labels are available but instance-level annotations are costly. Attention-based MIL methods learn to aggregate instance features by assigning weights that both drive predictions and serve as a proxy for instance importance~\cite{ilse_attention-based_2018}, making them attractive for clinical pipelines that require some degree of interpretability.
A well-documented failure mode of these methods is attention
concentration: the learned weights collapse onto a handful of
instances, creating attention sinks that yield overconfident
predictions and degrade generalization across cohorts and imaging sources~\cite{zhang_attention-challenging_2024}.
Existing remedies operate at the training level (attention masking~\cite{zhang_attention-challenging_2024}, self-supervised
pretraining~\cite{li_dual-stream_2021}, or teacher-student distillation~\cite{qu_bi-directional_nodate,tang_multiple_2023}). While effective, they often introduce additional stages or auxiliary losses that complicate the overall pipeline. 
Rather than regularizing attention after the fact, we redesign the aggregation module itself so that concentrated attention is structurally discouraged. Our initial observation is that attention concentration in MIL closely resembles the information collapse phenomenon recently investigated in large-scale language models~\cite{qiu_gated_2025}, where gating and normalization inside the attention block have proven effective countermeasures. 
Therefore, we translate these ideas into a compact MIL aggregator QG-MIL (Qwen Gated Multiple Instance Learning) that slots into any existing pipeline as a drop-in replacement in standard MIL setups, with no extra training stages or loss terms. 
A key point of this work is domain agnosticism. MIL problems in medical imaging range from gigapixel whole-slide images with thousands of patches to blood-smear analysis with hundreds of single-cell instances. An aggregation module that works well in one regime but not the other is of limited practical value. Therefore, we evaluate QG-MIL on both pathology and hematology benchmarks, using the same architecture and hyperparameters throughout, to stress-test
generalization across fundamentally different bag sizes and imaging modalities.
Our main contributions are:
\begin{itemize}
    \item We propose QG-MIL, a gated transformer aggregation
          module for MIL that mitigates attention concentration through
          architectural design alone, without auxiliary losses or
          multi-stage training, along with implementation code.
    \item We evaluate QG-MIL on six benchmarks across pathology and
          hematology, showing consistent improvements in predictive
          performance, attention distribution, and localization
          quality with the QG-MIL model and its ablations.
    \item We provide ablation studies isolating each design choice and show the impact of the cohort size and model depth on ablation performances.
\end{itemize}

\section{Methodology}
In MIL, each sample is a bag $\mathcal{B}=\{x_i\}_{i=1}^N$ with instance features $x_i\in\mathbb{R}^d$ and a single bag label $y$. QG-MIL maps instances in a bag to a shared latent space, processes them with $L$ stacked gated transformer blocks, and aggregates them through gated attention pooling to obtain a bag representation for classification. We denote $W_p$ as the input projection; $Q$, $K$, and $V$ as the query, key, and value projections; $A$ as the attention weights; $Z$ as the attention output before gating; $O$ as the gated attention output; and $W_o$ as the output projection.

Instances are first projected to dimension $D$, yielding projected instance embeddings $h$
\begin{equation}
h_i=\mathrm{Dropout}\big(\mathrm{RMSNorm}(W_p x_i)\big), 
\qquad 
H=[h_1;\dots;h_N]\in\mathbb{R}^{N\times D}.
\end{equation}

For each block, linear projections produce $\mathcal{Q},\mathcal{K},\mathcal{V}$, which are reshaped into multihead form with per-head dimension $d_h$ such that $D=H d_h$. Queries and keys are normalized per head for stability ($\widehat{Q}\ ,\widehat{K}$), and scaled dot-product attention is computed as

\begin{equation}
S=\frac{\widehat{Q}\,\widehat{K}^{\top}}{\sqrt{d_h}}, 
\qquad 
A=\mathrm{softmax}(S).
\end{equation}

The attention output of head $h$ for token $i$ is $Z_{i,h,:}=A_{i,h}V_{\cdot,h,:}\in\mathbb{R}^{d_h}$.
Gating is applied to $Z$ before the output projection. Crucially, the gate activations are computed via dedicated learnable linear projections parameterized by weights $W_{g,h}$. 
In the headwise formulation, the gate is computed via a projection to a scalar:
\begin{equation}
z_{i,h} = W_{g,h} Z_{i,h,:}, \qquad g_{i,h}=\sigma(z_{i,h})
\end{equation}
which modulates all features uniformly: $O_{i,h,:} = g_{i,h} Z_{i,h,:}$.
In the elementwise formulation, the projection maintains the feature dimension $d_h$, 
modulating each feature independently: $O_{i,h,:} = g_{i,h,:} \odot Z_{i,h,:}$.
Concatenating heads and applying the output matrix yields the block attention output, each block follows a pre-norm residual layout with a SwiGLU feed-forward network:

\begin{align}
\mathrm{Attn}(H) &= W_o\big(\mathrm{concat}_h O_{:,h,:}\big), &
\mathrm{FFN}(x)  &= W_d\big(\phi(W_g x)\odot W_u x\big).
\end{align}

and updates
\begin{align}
x' &= x + \mathrm{Dropout}\big(\mathrm{Attn}(\mathrm{Norm}(x))\big), \quad
x'' = x' + \mathrm{Dropout}\big(\mathrm{FFN}(\mathrm{Norm}(x'))\big).
\end{align}

Stacking $L$ such blocks yields refined instance embeddings.

Finally, gated attention pooling aggregates the final processed instances $h_i$ into a bag representation,
\begin{equation}
a_i = w^\top\big(\tanh(W_v h_i)\odot\sigma(W_u h_i)\big), 
\qquad
\alpha_i=\frac{\exp(a_i)}{\sum_j\exp(a_j)}, 
\qquad
z=\sum_i \alpha_i h_i,
\end{equation}
and $z$ is fed to a classification head to predict the bag label.
\\
Based on the architecture defined above, we evaluated several ablation variants to identify potential improvements and bottlenecks: \textit{QG-MIL Elementwise}, where gates are applied per feature dimension within each head; \textit{QG-MIL noGate}, where attention-output gating is removed; \textit{QG-MIL noQKnorm}, where per-head Q/K normalization is disabled; \textit{QG-MIL LayerNorm}, where RMSNorm is replaced with LayerNorm; \textit{QG-MIL Light}, which uses a reduced hidden dimension of 256; and \textit{QG-MIL Deep}, which consists of four stacked layers. The standard QG-MIL model contains 9 million parameters, while the Light variant is reduced to 2.4 million.


\noindent\textbf{Evaluation Data: }
For histopathology, we utilize two diagnostic whole-slide datasets: MSK~\cite{breast}(Breast): 130 WSIs from 78 patients, 36 with metastatic carcinoma, and LungHist700~\cite{LungHist700}(Lung): 691 images from 45 patients across normal and carcinoma classes.
Additionally, we include a prognostic benchmark: Prostate Cancer~\cite{ProstateBiopsy}(Prostate). This dataset comprises 587 biopsies from 213 initially benign patients, focusing on predicting future cancer development, cancer-free $\ge$ 8 years versus prostate cancer within 30 months.
To assess generalizability, the Breast and Lung datasets are encoded with foundation models: UNI2-h~\cite{uni}, Prov-Gigapath~\cite{gigapath}, and CONCHv1.5~\cite{conch}. The challenging prognostic prostate dataset is additionally processed with UNIv1~\cite{uni} and ResNet50~\cite{resnet50} to evaluate QG-MIL across both foundation and conventional CNN extractors.
For cell-level hematology, we evaluate on three benchmark datasets: \textit{AML-Hehr}~\cite{aml-hehr} 129 patients distributed in 5 classes, \textit{APL-AML}~\cite{aplaml} 106 patients grouped in 2 classes, and \textit{cAItomorph}~\cite{caitomorph} 2043 patients across 8 classes. These are processed using DinoBloom~\cite{DinoBloom}, a white blood cell specialized foundation model, across two encoder sizes, specifically Small and Large. All patch and cell encoders are frozen and used as static feature extractors.

\section{Experiments and Results}

\noindent\textbf{Experimental protocol.} 
To ensure robust evaluation across varying cohort sizes, we allocate a patient-stratified 20\% hold-out test set for final assessment. Within the remaining 80\% of the patient cohort, we employ a 5-fold cross-validation strategy to train five independent models. During inference, the predictions from these fold-specific models are ensembled on the fixed test set using mean probability aggregation.
Our data splitting and training regimen are explicitly designed to promote fair evaluation by adopting a shared scenario that facilitates direct comparisons with existing literature. By utilizing these common benchmark settings whenever applicable, we strictly adhere to the experimental parameters and patient splits established by previous works to ensure the reproducibility of our findings. Demonstrating this commitment to a shared evaluation framework, we utilize the original data splits defined by~\cite{caitomorph} for all hematology datasets. Finally, to quantify model stability, we report the standard deviation of the test-set performance across the five-fold-specific models.

\begin{table}[ht]
\centering
\caption{Summary classification performance on Breast and Lung pathology benchmarks. QG-MIL consistently improves F1. Best results per column are bolded; second-best are underlined.}
\label{tab:benchmark_lung_breast}
\resizebox{0.99\textwidth}{!}{
\begin{tabular}{l|ccc@{\hspace{0.2cm}}c|ccc@{\hspace{0.2cm}}c}
\hline
\textbf{Model} & \multicolumn{4}{c|}{\textbf{Breast}} & \multicolumn{4}{c}{\textbf{Lung}} \\ \hline
 & \textbf{UNI2-h} & \textbf{Prov-Gigapath} & \textbf{CONCHv1.5} & \textbf{Mean} & \textbf{UNI2-h} & \textbf{Prov-Gigapath} & \textbf{CONCHv1.5} & \textbf{Mean} \\ \hline
ABMIL & 81.21 {\scriptsize $\pm$  3.3} & 81.01 {\scriptsize $\pm$  5.9} & 91.30 {\scriptsize $\pm$  5.4} & 84.51 & 90.80 {\scriptsize $\pm$  3.1} & 85.56 {\scriptsize $\pm$  4.0} & 86.72 {\scriptsize $\pm$  2.0} & 87.69 \\
CLAM & 81.21 {\scriptsize $\pm$  3.3} & 79.72 {\scriptsize $\pm$  4.1} & 91.30 {\scriptsize $\pm$  5.4} & 84.08 & 89.94 {\scriptsize $\pm$  3.2} & 84.12 {\scriptsize $\pm$  3.87} & 85.03 {\scriptsize $\pm$  2.6} & 86.36 \\
DSMIL & 82.50 {\scriptsize $\pm$  4.9} & 81.13 {\scriptsize $\pm$  18.2} & 93.73 {\scriptsize $\pm$  2.6} & 85.79 & 91.11 {\scriptsize $\pm$  3.1} & 84.45 {\scriptsize $\pm$  3.2} & 84.48 {\scriptsize $\pm$  3.9} & 86.68 \\
ILRA & 75.99 {\scriptsize $\pm$  10.0} & 81.16 {\scriptsize $\pm$  9.4} & 87.72 {\scriptsize $\pm$  5.1} & 81.62 & 88.74 {\scriptsize $\pm$  3.7} & 80.70 {\scriptsize $\pm$  3.5} & 86.34 {\scriptsize $\pm$  1.9} & 85.26 \\
RRT & 82.50 {\scriptsize $\pm$  4.9} & 81.01 {\scriptsize $\pm$  5.9} & 94.87 {\scriptsize $\pm$  0.0} & 86.13 & 92.17 {\scriptsize $\pm$  2.1} & 86.63 {\scriptsize $\pm$  2.70} & 84.68 {\scriptsize $\pm$  4.71} & 87.83 \\
Transformer & 77.95 {\scriptsize $\pm$  7.2} & 85.09 {\scriptsize $\pm$  6.2} & 91.30 {\scriptsize $\pm$  5.4} & 84.78 & 90.63 {\scriptsize $\pm$  2.6} & 86.79 {\scriptsize $\pm$  2.9} & 85.02 {\scriptsize $\pm$  4.2} & 87.48 \\
TransMIL & 81.21 {\scriptsize $\pm$  3.3} & 79.24 {\scriptsize $\pm$  8.7} & 90.01 {\scriptsize $\pm$  6.7} & 83.49 & 90.58 {\scriptsize $\pm$  3.2} & 86.54 {\scriptsize $\pm$  4.5} & 83.83 {\scriptsize $\pm$  3.7} & 86.98 \\
WIKG & 82.71 {\scriptsize $\pm$  0.0} & \textbf{91.51 {\scriptsize $\pm$  5.3}} & 91.30 {\scriptsize $\pm$  5.4} & 88.51 & 91.95 {\scriptsize $\pm$  4.3} & 84.75 {\scriptsize $\pm$  3.0} & 87.91 {\scriptsize $\pm$  3.0} & 88.20 \\ 
MeanMIL & 53.66 {\scriptsize $\pm$ 13.0}  & 53.82 {\scriptsize $\pm$ 12.5}& 41.96 {\scriptsize $\pm$ 0.6} & 49.81 & 86.28 {\scriptsize $\pm$ 15.0}& 73.77 {\scriptsize $\pm$ 13.9} & 85.25 {\scriptsize $\pm$ 9.4}  & 81.77 \\
\hline
QG-MIL & \textbf{85.29 {\scriptsize $\pm$  3.5}} & 87.72 {\scriptsize $\pm$  5.1} & 91.30 {\scriptsize $\pm$  5.4} & 88.10 & \underline{93.00 {\scriptsize $\pm$  3.5}} & 87.39 {\scriptsize $\pm$  1.9} & 87.45 {\scriptsize $\pm$  3.2} & \underline{89.28} \\
QG-MIL Deep & \textbf{85.29 {\scriptsize $\pm$  3.5}} & 87.52 {\scriptsize $\pm$  7.3} & \textbf{98.97 {\scriptsize $\pm$  2.3}} & \textbf{90.59} & 92.65 {\scriptsize $\pm$  1.8} & 86.10 {\scriptsize $\pm$  2.5} & 85.52 {\scriptsize $\pm$  3.1} & 88.09 \\
QG-MIL Elementwise& 82.30 {\scriptsize $\pm$  7.0} & 90.16 {\scriptsize $\pm$  5.1} & 91.30 {\scriptsize $\pm$  5.4} & 87.92 & 92.70 {\scriptsize $\pm$  2.0} & 86.14 {\scriptsize $\pm$  3.4} & \underline{87.98 {\scriptsize $\pm$  2.7}} & 88.94 \\
QG-MIL Layernorm & \underline{84.00 {\scriptsize $\pm$  2.9}} & \underline{91.45 {\scriptsize $\pm$  3.1}} & 95.78 {\scriptsize $\pm$  4.5} & \underline{90.41} & 92.67 {\scriptsize $\pm$  3.3} & 87.52 {\scriptsize $\pm$  2.6} & 85.89 {\scriptsize $\pm$  4.2} & 88.69 \\
QG-MIL Light & 81.21 {\scriptsize $\pm$  3.3} & 89.02 {\scriptsize $\pm$  4.3} & \underline{95.90 {\scriptsize $\pm$  2.3}} & 88.71 & 90.30 {\scriptsize $\pm$  8.7} & 86.29 {\scriptsize $\pm$  6.1} & 86.27 {\scriptsize $\pm$  1.8} & 87.62 \\
QG-MIL no Gate & \textbf{85.29 {\scriptsize $\pm$  3.5}} & 86.23 {\scriptsize $\pm$  7.5} & \underline{95.90 {\scriptsize $\pm$  2.3}} & 89.14 & \textbf{93.34 {\scriptsize $\pm$  3.7}} & \textbf{87.78 {\scriptsize $\pm$  4.5}} & \textbf{88.30 {\scriptsize $\pm$  4.9}} & \textbf{89.81} \\
QG-MIL no QKnorm & \underline{84.00 {\scriptsize $\pm$  2.9}} & \underline{91.45 {\scriptsize $\pm$  3.1}} & 95.78 {\scriptsize $\pm$  4.5} & \underline{90.41} & 92.09 {\scriptsize $\pm$ 1.9} & \underline{87.54 {\scriptsize $\pm$  3.8}} & 85.89 {\scriptsize $\pm$  4.2} & 88.51 \\
\hline
Mean Baselines & 77.66 & 79.30 & 85.94 & 80.97 & 90.24 & 83.70 & 85.47 & 86.47 \\
Mean QG-MIL & \textbf{83.91} & \textbf{89.08} & \textbf{94.99} & \textbf{89.33} & \textbf{92.39} & \textbf{86.97} & \textbf{86.76} & \textbf{88.71} \\
\hline
\end{tabular}
}
\end{table}

Given the pronounced class imbalance typical of medical datasets, we select macro F1 as the primary evaluation metric. All models are trained using CrossEntropy loss for a maximum of 150 epochs, with early stopping applied after 10 epochs without improvement in validation loss, dropout of 0.25, and embedding size $D$ of 512. Model selection for each fold is based exclusively on the lowest validation loss. To maintain architectural consistency and isolate the contribution of the proposed design, all QG-MIL variants are implemented with two layers, matching the depth of baseline MIL aggregators, including ILRA, RRT, Transformer, and TransMIL, following the standard implementation of \cite{shao2025do}. While further parameter tuning could potentially improve absolute performance, exhaustive optimization is beyond the scope of this study. Our goal is instead to assess architectural behavior under controlled and comparable conditions.

\begin{table}[ht]
\centering
\caption{Classification performance on the Prostate Cancer prognostic task. The QG-MIL LayerNorm variant achieves the highest overall average (57.55\%), demonstrating superior early prognostic capability compared to baseline aggregators. Best results per column are bolded; second-best are underlined.}
\label{tab:benchmark_prostate}
\resizebox{0.99\textwidth}{!}{
\begin{tabular}{l|cccccc}
\hline
\textbf{Model} & \multicolumn{6}{c}{\textbf{Prostate}} \\
\hline
 & \textbf{CONCHv1.5} & \textbf{Prov-Gigapath} & \textbf{ResNet50} & \textbf{UNIv1} & \textbf{UNI2-h} & \textbf{Mean} \\
\hline
ABIMIL & 56.09 {\scriptsize $\pm$ 3.0} & 56.09 {\scriptsize $\pm$ 4.8} & 44.16 {\scriptsize $\pm$ 0.0} & \underline{59.68 {\scriptsize $\pm$ 4.9}} & 56.09 {\scriptsize $\pm$ 2.5} & 54.42 \\
CLAM & 51.80 {\scriptsize $\pm$ 6.9} & 56.30 {\scriptsize $\pm$ 1.4} & 44.10 {\scriptsize $\pm$ 0.0} & 57.20 {\scriptsize $\pm$ 4.7} & 57.50 {\scriptsize $\pm$ 2.3} & 53.38 \\
DSMIL & \textbf{59.10 {\scriptsize $\pm$ 7.6}} & \underline{58.00 {\scriptsize $\pm$ 2.6}} & 49.90 {\scriptsize $\pm$ 4.7} & 59.80 {\scriptsize $\pm$ 3.2} & 53.90 {\scriptsize $\pm$ 4.2} & 56.14 \\
ILRA & 52.75 {\scriptsize $\pm$ 5.0} & 56.09 {\scriptsize $\pm$ 4.5} & 54.39 {\scriptsize $\pm$ 2.3} & 59.68 {\scriptsize $\pm$ 2.2} & 54.39 {\scriptsize $\pm$ 3.8} & 55.46 \\
RRT & 56.09 {\scriptsize $\pm$ 4.3} & 49.33 {\scriptsize $\pm$ 5.8} & 50.82 {\scriptsize $\pm$ 3.6} & 54.39 {\scriptsize $\pm$ 5.3} & 54.39 {\scriptsize $\pm$ 2.9} & 53.00 \\
Transformer & 54.39 {\scriptsize $\pm$ 7.6} & \textbf{60.91 {\scriptsize $\pm$ 3.2}} & \underline{56.09 {\scriptsize $\pm$ 5.6}} & 59.05 {\scriptsize $\pm$ 5.1} & 54.39 {\scriptsize $\pm$ 2.5} & \underline{56.97} \\
TransMIL & 54.39 {\scriptsize $\pm$ 7.5} & 54.39 {\scriptsize $\pm$ 4.2} & \underline{56.09 {\scriptsize $\pm$ 4.4}} & 56.09 {\scriptsize $\pm$ 4.0} & 56.09 {\scriptsize $\pm$ 5.6} & 55.41 \\
WIKG & 54.39 {\scriptsize $\pm$ 2.9} & 54.39 {\scriptsize $\pm$ 3.7} & \underline{56.09 {\scriptsize $\pm$ 3.9}} & 54.39 {\scriptsize $\pm$ 5.3} & 52.75 {\scriptsize $\pm$ 6.1} & 54.40 \\
MeanMIL & 44.16 {\scriptsize $\pm$ 0.0} & 47.17 {\scriptsize $\pm$ 6.7} & 44.16 {\scriptsize $\pm$ 0.0} & 50.49 {\scriptsize $\pm$ 6.6} & 54.07 {\scriptsize $\pm$ 6.0} & 48.01 \\
\hline
QG-MIL & 56.09 {\scriptsize $\pm$ 1.5} & 54.39 {\scriptsize $\pm$ 3.5} & 52.75 {\scriptsize $\pm$ 5.7} & 56.09 {\scriptsize $\pm$ 1.6} & \textbf{59.68 {\scriptsize $\pm$ 1.9}} & 55.80 \\
QG-MIL Deep & 56.09 {\scriptsize $\pm$ 1.7} & 54.39 {\scriptsize $\pm$ 2.5} & 49.33 {\scriptsize $\pm$ 5.8} & 56.09 {\scriptsize $\pm$ 5.4} & \underline{57.84 {\scriptsize $\pm$ 3.2}} & 54.75 \\
QG-MIL Elementwise& 56.09 {\scriptsize $\pm$ 1.5} & 57.84 {\scriptsize $\pm$ 3.6} & 54.39 {\scriptsize $\pm$ 2.6} & 57.84 {\scriptsize $\pm$ 3.9} & 56.09 {\scriptsize $\pm$ 3.1} & 56.45 \\
QG-MIL Layernorm & \underline{57.84 {\scriptsize $\pm$ 2.2}} & 54.39 {\scriptsize $\pm$ 4.1} & \textbf{57.84 {\scriptsize $\pm$ 3.7}} & \textbf{61.61 {\scriptsize $\pm$ 5.2}} & 56.09 {\scriptsize $\pm$ 3.0} & \textbf{57.55} \\
QG-MIL Light & 54.39 {\scriptsize $\pm$ 3.7} & 54.39 {\scriptsize $\pm$ 3.8} & 51.14 {\scriptsize $\pm$ 2.2} & 57.84 {\scriptsize $\pm$ 2.2} & 56.09 {\scriptsize $\pm$ 2.8} & 54.77 \\
QG-MIL no Gate & 47.51 {\scriptsize $\pm$ 4.8} & 54.39 {\scriptsize $\pm$ 2.1} & 49.33 {\scriptsize $\pm$ 2.9} & 56.09 {\scriptsize $\pm$ 5.1} & \underline{57.84 {\scriptsize $\pm$ 2.6}} & 53.03 \\
QG-MIL no QKnorm & 56.09 {\scriptsize $\pm$ 2.3} & 54.39 {\scriptsize $\pm$ 1.9} & \textbf{57.84 {\scriptsize $\pm$ 3.7}} & \underline{59.68 {\scriptsize $\pm$ 3.8}} & 56.09 {\scriptsize $\pm$ 3.5} & 56.82 \\
\hline
Mean Baselines & 53.68 & 54.74 & 50.64 & 56.75 & 54.84 & 54.13 \\
Mean QG-MIL & \textbf{54.87} & \textbf{54.88} & \textbf{53.23} & \textbf{57.89} & \textbf{57.10} & \textbf{55.59} \\
\hline
\end{tabular}
}
\end{table}

\noindent\textbf{General Architectural and Performance Findings.}
Across both computational pathology and hematology domains, consistent patterns emerge regarding the behavior and efficacy of the proposed QG-MIL framework. First, we observe that the optimal architectural complexity is strictly dictated by cohort size. In low-patient settings, stringent gating mechanisms and Q/K normalization introduce variance that degrades generalization, indicating that smaller datasets achieve optimal performance when regularization constraints are relaxed. Conversely, medium-to-large cohorts remain highly stable during optimization and benefit significantly from the full gating architecture. 

Second, QG-MIL exhibits a strong positive correlation between network depth and predictive performance. Assuming sufficient data, deeper aggregation variants, such as QG-MIL Deep, consistently unlock superior macro F1 scores, demonstrating that the architecture scales effectively without suffering from severe optimization bottlenecks. Finally, the refined aggregation mechanisms prove exceptionally adept at capturing subtle morphological features. This capability drives consistent performance improvements over baseline MIL methods, such as standard Transformers and WIKG, and becomes distinctly advantageous in highly constrained, early-stage prognostic scenarios.

\noindent\textbf{Pathology Benchmarks on Diagnosis.}
The diagnostic benchmarks contrast a low-patient Lung dataset with a medium-to-large Breast dataset. Reflecting our general findings, the Lung benchmark favors relaxed configurations, achieving a peak macro F1 of $93.3\%$. This establishes a substantial performance margin over previous studies, which reported classification accuracies of $81.6\%$ \cite{yixuan_cost-sensitive_nodate} and $81.0\%$ \cite{LungHist700}. In contrast, the larger Breast cohort leverages deeper aggregation effectively. The QG-MIL Deep variant achieves the highest macro F1 of $98.9\%$. Across all encoders and folds, the QG-MIL variants consistently achieve higher mean macro-F1 scores, with only WIKG showing comparable average baseline performance. Full results are reported in \Cref{tab:benchmark_lung_breast}.

\noindent\textbf{Pathology Benchmarks on Prognosis.}
The Prostate dataset introduces a highly challenging prognostic task: predicting the future onset of malignancy in patients initially diagnosed as strictly benign. Due to the inherent difficulty of early prediction, overall metrics are markedly lower across all models. While baseline methods achieve an average macro F1 score of $54.1\%$ (peaking at $56.9\%$ for the standard Transformer), the QG-MIL Layernorm variant achieves the highest overall average macro F1 of $57.5\%$. This highlights the framework's superior ability to isolate the elusive, early-stage morphological signals required to predict malignant transformations prior to standard clinical diagnosis (\Cref{tab:benchmark_prostate}).

\noindent\textbf{Hematology Benchmarks on Diagnosis.}
Beyond solid tissue pathology, QG-MIL demonstrates robust efficacy in hematological tasks, outperforming baseline methods by an average of over $5\%$ in macro F1 score. For the classification of Acute Promyelocytic Leukemia (APL) versus non-APL, the model achieves a strong macro F1 of $69.5\%$ utilizing the deep configuration, directly mirroring the depth-to-regularization trend observed in the Lung dataset. In Acute Myeloid Leukemia (AML) subtyping, QG-MIL reaches a macro F1 of $86.0\%$, surpassing other specialized MIL methods \cite{hehr_explainable_nodate} by $5\%$. Furthermore, on the cAItomorph dataset, the QG-MIL Deep variant achieves a maximum weighted F1 of $67.9\%$, representing a $3.7\%$ performance increase over previous studies evaluating up to $8$-layer transformer models \cite{caitomorph}. Full results are reported in \Cref{tab:benchmark_blood}.

\begin{table}[htb]
\centering
\caption{Classification performance macro F1 score across cell-level hematology benchmarks. Results demonstrate the domain-agnostic scalability of QG-MIL, as it outperforms established baselines by an average of >5\% across varying hematological cell analysis tasks. Best results per column are bolded; second-best are underlined.}

\label{tab:benchmark_blood}
\resizebox{0.99\textwidth}{!}{
\begin{tabular}{l|ccc|ccc|ccc}
\hline
\textbf{Model}
  & \multicolumn{3}{c|}{\textbf{AML-Hehr}}
  & \multicolumn{3}{c|}{\textbf{APL-AML}}
  & \multicolumn{3}{c}{\textbf{cAItomorph}} \\
 & \textbf{Bloom-S} & \textbf{Bloom-L} & \textbf{Mean}
 & \textbf{Bloom-S} & \textbf{Bloom-L} & \textbf{Mean}
 & \textbf{Bloom-S} & \textbf{Bloom-L} & \textbf{Mean} \\
\hline
ABMIL & 67.79 \scriptsize{$\pm$ 6.8} & 69.81 \scriptsize{$\pm$ 8.5} & 68.80
      & 59.01 \scriptsize{$\pm$ 6.8} & 39.50 \scriptsize{$\pm$ 14.1} & 49.26
      & 57.23 \scriptsize{$\pm$ 2.7} & 55.40 \scriptsize{$\pm$ 3.2} & 56.32 \\
TransMIL & 72.59 \scriptsize{$\pm$ 8.1} & 64.86 \scriptsize{$\pm$ 9.7} & 68.72
         & 39.50 \scriptsize{$\pm$ 8.6} & 39.50 \scriptsize{$\pm$ 9.4} & 39.50
         & 58.24 \scriptsize{$\pm$ 3.9} & 57.83 \scriptsize{$\pm$ 1.6} & 58.04 \\
Transformer & 76.97 \scriptsize{$\pm$ 9.1} & 81.14 \scriptsize{$\pm$ 8.0} & 79.06
            & \underline{64.44 \scriptsize{$\pm$ 6.8}} & 64.44 \scriptsize{$\pm$ 3.4} & 64.44
            & \underline{66.39 \scriptsize{$\pm$ 2.3}} & 63.45 \scriptsize{$\pm$ 1.5} & 64.92 \\
DSMIL & 63.11 \scriptsize{$\pm$ 10.1} & 73.14 \scriptsize{$\pm$ 4.0} & 68.12
      & 41.98 \scriptsize{$\pm$ 3.3} & 49.58 \scriptsize{$\pm$ 9.6} & 45.78
      & 52.02 \scriptsize{$\pm$ 3.8} & 51.76 \scriptsize{$\pm$ 2.9} & 51.89 \\
DFTD & 64.03 \scriptsize{$\pm$ 7.1} & 73.50 \scriptsize{$\pm$ 12.0} & 68.76
     & \underline{64.44 \scriptsize{$\pm$ 8.7}} & \underline{68.12 \scriptsize{$\pm$ 16.2}} & 66.28
     & 54.60 \scriptsize{$\pm$ 2.0} & 60.19 \scriptsize{$\pm$ 1.1} & 57.40 \\
WIKG & 57.81 \scriptsize{$\pm$ 11.5} & 66.60 \scriptsize{$\pm$ 8.9} & 62.20
     & 55.56 \scriptsize{$\pm$ 6.7} & 41.98 \scriptsize{$\pm$ 10.1} & 48.77
     & 60.56 \scriptsize{$\pm$ 4.0} & 61.52 \scriptsize{$\pm$ 2.1} & 61.04 \\
ILRA & 70.33 \scriptsize{$\pm$ 7.1} & 68.82 \scriptsize{$\pm$ 5.8} & 69.57
     & 59.01 \scriptsize{$\pm$ 11.5} & \underline{68.12 \scriptsize{$\pm$ 4.5}} & 63.56
     & 63.80 \scriptsize{$\pm$ 2.5} & 60.86 \scriptsize{$\pm$ 1.2} & 62.33 \\
RRT & 68.56 \scriptsize{$\pm$ 4.1} & 62.80 \scriptsize{$\pm$ 6.7} & 65.68
    & 59.01 \scriptsize{$\pm$ 6.5} & 59.01 \scriptsize{$\pm$ 7.2} & 59.01
    & 64.78 \scriptsize{$\pm$ 2.6} & 63.19 \scriptsize{$\pm$ 2.3} & 63.98 \\
CLAM & 75.75 \scriptsize{$\pm$ 4.6} & 77.17 \scriptsize{$\pm$ 8.7} & 76.46
     & 46.67 \scriptsize{$\pm$ 3.2} & 46.67 \scriptsize{$\pm$ 10.2} & 46.67
     & 57.43 \scriptsize{$\pm$ 2.6} & 57.94 \scriptsize{$\pm$ 2.7} & 57.68 \\
MeanMIL & 66.55 \scriptsize{$\pm$ 4.8} & 69.15 \scriptsize{$\pm$ 7.5} & 67.85
          & 53.12 \scriptsize{$\pm$ 3.2} & 59.01 \scriptsize{$\pm$ 16.8} & 56.07
          & 58.56 \scriptsize{$\pm$ 2.7} & 60.06 \scriptsize{$\pm$ 1.6} & 59.31 \\
\hline
QG-MIL & 76.62 \scriptsize{$\pm$ 5.7} & 71.33 \scriptsize{$\pm$ 5.2} & 73.97
      & 59.01 \scriptsize{$\pm$ 14.7} & 59.01 \scriptsize{$\pm$ 11.6} & 59.01
      & 63.77 \scriptsize{$\pm$ 6.4} & \textbf{66.27 \scriptsize{$\pm$ 5.0}} & 65.02 \\
QG-MIL Elementwise& 79.98 \scriptsize{$\pm$ 6.1} & \underline{82.78 \scriptsize{$\pm$ 4.8}} & \underline{81.38}
                   & 59.01 \scriptsize{$\pm$ 8.9} & 53.12 \scriptsize{$\pm$ 15.6} & 56.06
                   & 65.20 \scriptsize{$\pm$ 7.7} & 65.81 \scriptsize{$\pm$ 6.2} & \underline{65.50} \\
QG-MIL no Gate & 79.44 \scriptsize{$\pm$ 4.2} & 79.69 \scriptsize{$\pm$ 8.8} & 79.56
                & \underline{64.44 \scriptsize{$\pm$ 11.4}} & \textbf{69.51 \scriptsize{$\pm$ 5.8}} & \underline{66.97}
                & 64.71 \scriptsize{$\pm$ 5.2} & \underline{66.13 \scriptsize{$\pm$ 6.1}} & 65.42 \\
QG-MIL no QKnorm & 79.98 \scriptsize{$\pm$ 7.8} & 75.93 \scriptsize{$\pm$ 8.0} & 77.96
                  & \textbf{69.51 \scriptsize{$\pm$ 14.3}} & 64.44 \scriptsize{$\pm$ 10.0} & \underline{66.97}
                  & 61.63 \scriptsize{$\pm$ 9.0} & 65.82 \scriptsize{$\pm$ 5.2} & 63.72 \\
QG-MIL Layernorm & \underline{82.67 \scriptsize{$\pm$ 7.8}} & 75.93 \scriptsize{$\pm$ 7.3} & 79.30
                 & 59.01 \scriptsize{$\pm$ 2.4} & 64.44 \scriptsize{$\pm$ 12.5} & 61.72
                 & 64.53 \scriptsize{$\pm$ 10.5} & 62.53 \scriptsize{$\pm$ 5.2} & 63.53 \\
QG-MIL Deep & 79.98 \scriptsize{$\pm$ 7.0} & \textbf{86.02 \scriptsize{$\pm$ 4.7}} & \textbf{83.00}
            & \textbf{69.51 \scriptsize{$\pm$ 5.1}} & \textbf{69.51 \scriptsize{$\pm$ 4.4}} & \textbf{69.51}
            & \textbf{67.97 \scriptsize{$\pm$ 6.1}} & 63.80 \scriptsize{$\pm$ 9.1} & \textbf{65.88} \\
QG-MIL Light & \textbf{82.90 \scriptsize{$\pm$ 8.3}} & 77.52 \scriptsize{$\pm$ 4.1} & 80.21
             & 60.80 \scriptsize{$\pm$ 12.5} & 64.44 \scriptsize{$\pm$ 11.0} & 62.62
             & 60.95 \scriptsize{$\pm$ 2.2} & 63.17 \scriptsize{$\pm$ 10.2} & 62.06 \\
\hline
Mean Baselines & 68.55 & 70.87 & 69.71
                 & 54.40 & 52.99 & 53.70
                 & 59.45 & 59.13 & 59.29 \\
\textbf{Mean QG-MIL} 
& \textbf{80.22} & \textbf{78.46} & \textbf{79.34}
& \textbf{63.04} & \textbf{63.50} & \textbf{63.27}
& \textbf{64.11} & \textbf{64.79} & \textbf{64.45} \\
\hline
\end{tabular}
}
\end{table}
\begin{figure}[htb]
    \centering
    \includegraphics[width=0.99\linewidth]{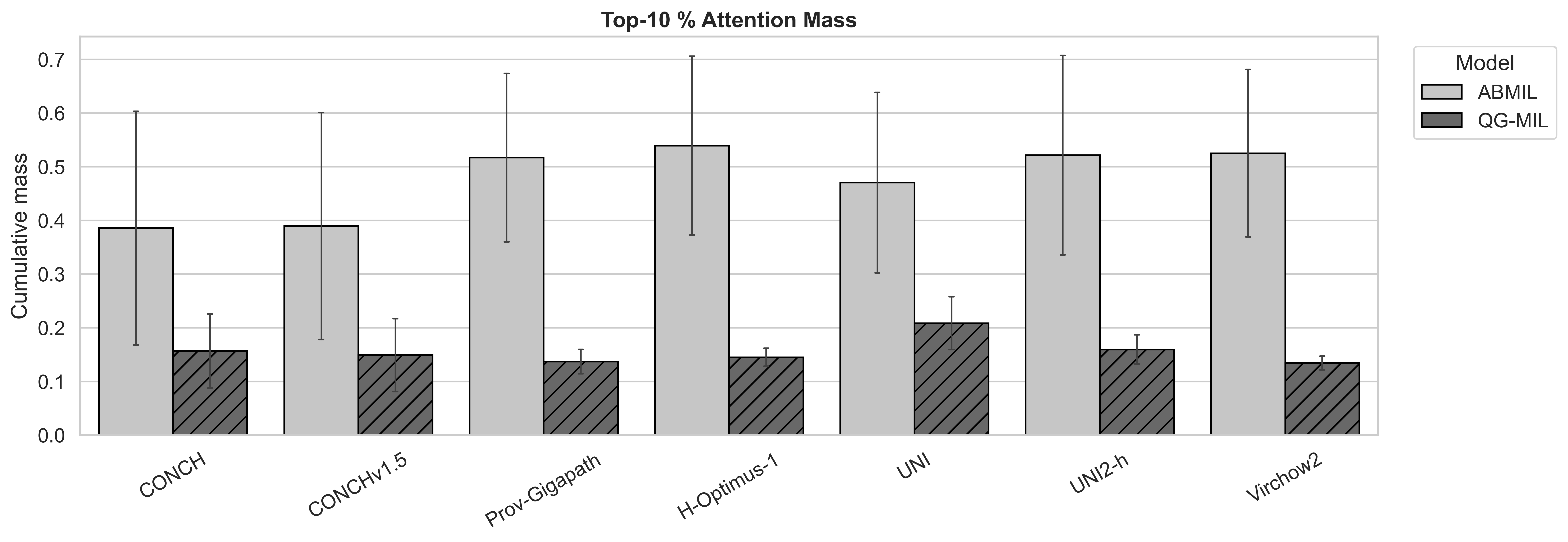}
    \caption{Mean top-10 attention mass extracted from QG-MIL and ABMIL models on the Breast dataset test set. Lower top-k mass indicates more distributed attention and reduced attention sink. QG-MIL variants show reduced concentration compared to ABMIL, suggesting improved instance coverage. Error bars denote standard deviation across test set images.}
    \label{fig:topk_mass}
\end{figure}

\noindent\textbf{QG-MIL improves attention distribution.}
While the quantitative results underscore the strong predictive performance of the QG-MIL family, the primary motivation behind QG-MIL is to improve the distribution of attention weights in MIL and to enhance robustness against optimization instabilities.
\begin{figure}[ht]
    \centering
    \includegraphics[width=0.99\linewidth]{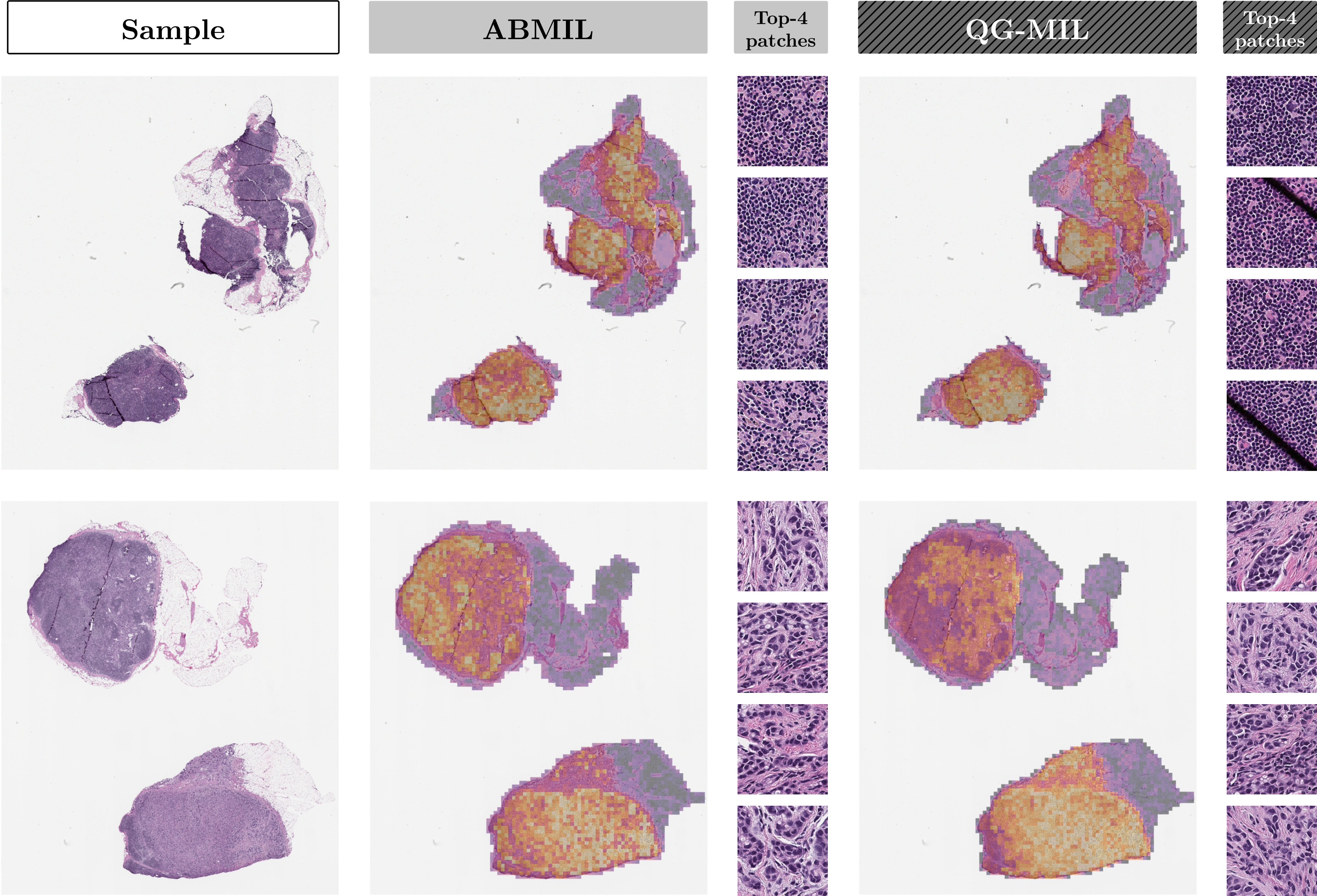}
    \caption{Qualitative attention overlays on example cases. Left: sample whole slide image. Middle: normalized attention heatmap from ABMIL and Top-4 patches based on attention score. Right: normalized attention heatmap from QG-MI along the Top-4 patches. QG-MIL highlights a visually smoother attention distribution, avoiding salt-and-pepper-like attention peaks for single patches.}
    \label{fig:visual_overlays}
\end{figure}
In \cref{fig:topk_mass}, we report the mean top-$10\%$ attention mass on the Breast dataset test set. QG-MIL maintains a near-constant mass of approximately $15\%$, effectively avoiding the attention-sink behavior seen in ABMIL, which concentrates up to $54\%$ of the total attention within the top $10\%$ of instances. 
To rigorously quantify this mitigation, we evaluate the attention distribution using the Gini index and entropy, where lower Gini and higher entropy indicate uniformity. ABMIL exhibits highly concentrated attention (Gini: $0.62\pm0.13$, entropy: $0.90 \pm 0.05$). In contrast, QG-MIL achieves a significantly more distributed profile (Gini: $0.16\pm0.07$, entropy: $0.99 \pm 0.01$). This quantitatively supports that QG-MIL's internal architectural gating intrinsically regularizes the pooling distribution without requiring auxiliary entropy losses.
This distributed behavior is visually confirmed in \cref{fig:visual_overlays}. Compared to ABMIL, QG-MIL produces smoother, spatially distributed attention maps. Notably, the top-4 attended patches in QG-MIL focus on similar morphological structures rather than isolated high-magnitude peaks, suggesting improved instance coverage and reduced sensitivity to local attention maxima.

\section{Conclusion and Limitations}
\label{sec:conclusion}
\subsection{Conclusion and Limitations.}
We introduce QG-MIL, a domain-agnostic gated transformer aggregation module that structurally mitigates attention sinks in medical imaging MIL without auxiliary losses. Across six pathology and hematology benchmarks, QG-MIL outperformed leading baselines by an average of $+6.1$ macro F1 points and yielded a smoother, more clinically plausible attention distribution.
Despite these advances, our study presents notable limitations. First, the full gated architecture can introduce variance in small patient cohorts, where simpler QG-MIL variants without gating are more effective. Second, while fine-grained gating and SwiGLU modules improve performance, they increase training memory and time relative to standard aggregators. Importantly, this overhead does not translate to prohibitive inference costs. For realistic bag sizes of $512$ and $1024$, QG-MIL requires approximately $7$ and $18$ GFLOPs, respectively, and thus exhibits a lower computational burden than RRT and TransMIL. Finally, exhaustive hyperparameter optimization was omitted to ensure fair baseline comparisons. 
Future work will explore adaptive gating mechanisms that dynamically scale complexity based on cohort size, alongside expanding evaluations to multi-modal clinical pipelines to further validate QG-MIL's generalizability.
\noindent\textbf{CO2 Emissions from Experiments.}
Our experiments were run on our in-house infrastructure using the NVIDIA A100 PCIe 80 GB hardware; the total emissions are estimated at 7.78 kg CO2eq.

%
%
%
%
\bibliographystyle{unsrtnat}
\bibliography{bib}

\end{document}